%% file: main.tex
\documentclass[acmtog]{acmart}

\usepackage{graphicx}
\usepackage{amsmath}
\usepackage{booktabs}
\usepackage{caption}
\usepackage{subcaption}
\input{math_commands.tex}

\usepackage[capitalize,nameinlink]{cleveref}
\crefname{section}{Sec.}{Secs.}
\Crefname{section}{Section}{Sections}
\Crefname{table}{Table}{Tables}
\crefname{table}{Tab.}{Tabs.}

\usepackage{enumitem}
\usepackage{color}
\usepackage{xspace}
\makeatletter
\DeclareRobustCommand\onedot{\futurelet\@let@token\@onedot}
\def\@onedot{\ifx\@let@token.\else.\null\fi\xspace}

\makeatother

\AtBeginDocument{%
  }

\copyrightyear{2024}
\acmYear{2024}
\setcopyright{rightsretained}
\acmConference[SA Conference Papers '24]{SIGGRAPH Asia 2024 Conference Papers}{December 3--6, 2024}{Tokyo, Japan}
\acmBooktitle{SIGGRAPH Asia 2024 Conference Papers (SA Conference Papers '24), December 3--6, 2024, Tokyo, Japan}\acmDOI{10.1145/3680528.3687653}
\acmISBN{979-8-4007-1131-2/24/12}
\acmSubmissionID{796}

\begin{document}

\title{URAvatar: Universal Relightable Gaussian Codec Avatars}

\author{Junxuan Li}
\email{junxuanli@meta.com}
\affiliation{%
 \institution{Codec Avatars Lab, Meta}
 \city{Pittsburgh}
 \state{Pennsylvania}
 \country{USA}
}
\author{Chen Cao}
\email{chencao@meta.com}
\affiliation{%
 \institution{Codec Avatars Lab, Meta}
 \city{Pittsburgh}
 \state{Pennsylvania}
 \country{USA}
}

\author{Gabriel Schwartz}
\email{gbschwartz@meta.com}
\affiliation{%
 \institution{Codec Avatars Lab, Meta}
 \city{Pittsburgh}
 \state{Pennsylvania}
 \country{USA}
}
\author{Rawal Khirodkar}
\email{rawalk@meta.com}
\affiliation{%
 \institution{Codec Avatars Lab, Meta}
 \city{Pittsburgh}
 \state{Pennsylvania}
 \country{USA}
}
\author{Christian Richardt}
\email{crichardt@meta.com}
\affiliation{%
 \institution{Codec Avatars Lab, Meta}
 \city{Pittsburgh}
 \state{Pennsylvania}
 \country{USA}
}
\author{Tomas Simon}
\email{tsimon@meta.com}
\affiliation{%
 \institution{Codec Avatars Lab, Meta}
 \city{Pittsburgh}
 \state{Pennsylvania}
 \country{USA}
}
\author{Yaser Sheikh}
\email{yasers@meta.com}
\affiliation{%
 \institution{Codec Avatars Lab, Meta}
 \city{Pittsburgh}
 \state{Pennsylvania}
 \country{USA}
}
\author{Shunsuke Saito}
\email{shunsuke.saito16@gmail.com}

\affiliation{%
 \institution{Codec Avatars Lab, Meta}
 \city{Pittsburgh}
 \state{Pennsylvania}
 \country{USA}
}

% \affiliation{%
%  \institution{The Th{\o}rv{\"a}ld Group}


\begin{abstract}
We present a new approach to creating photorealistic and relightable head avatars from a phone scan with unknown illumination. The reconstructed avatars can be animated and relit in real time with the global illumination of diverse environments. Unlike existing approaches that estimate parametric reflectance parameters via inverse rendering, our approach directly models learnable radiance transfer that incorporates global light transport in an efficient manner for real-time rendering. However, learning such a complex light transport that can generalize across identities is non-trivial. A phone scan in a \emph{single} environment lacks sufficient information to infer how the head would appear in general environments. To address this, we build a universal relightable avatar model represented by 3D Gaussians. We train on hundreds of high-quality multi-view human scans with controllable point lights.
High-resolution geometric guidance further enhances the reconstruction accuracy and generalization. Once trained, we finetune the pretrained model on a phone scan using inverse rendering to obtain a personalized relightable avatar. Our experiments establish the efficacy of our design, outperforming existing approaches while retaining real-time rendering capability.
\end{abstract}

\begin{CCSXML}
	<ccs2012>
	<concept>
	<concept_id>10010147.10010178.10010224.10010245.10010254</concept_id>
	<concept_desc>Computing methodologies~Reconstruction</concept_desc>
	<concept_significance>500</concept_significance>
	</concept>
	<concept>
	<concept_id>10010147.10010371.10010352</concept_id>
	<concept_desc>Computing methodologies~Animation</concept_desc>
	<concept_significance>500</concept_significance>
	</concept>
	</ccs2012>
\end{CCSXML}

\ccsdesc[500]{Computing methodologies~Reconstruction}
\ccsdesc[500]{Computing methodologies~Animation}
\keywords{3D Avatar Creation, Neural Rendering}
\begin{teaserfigure}
\centering
  \includegraphics[width=\textwidth]{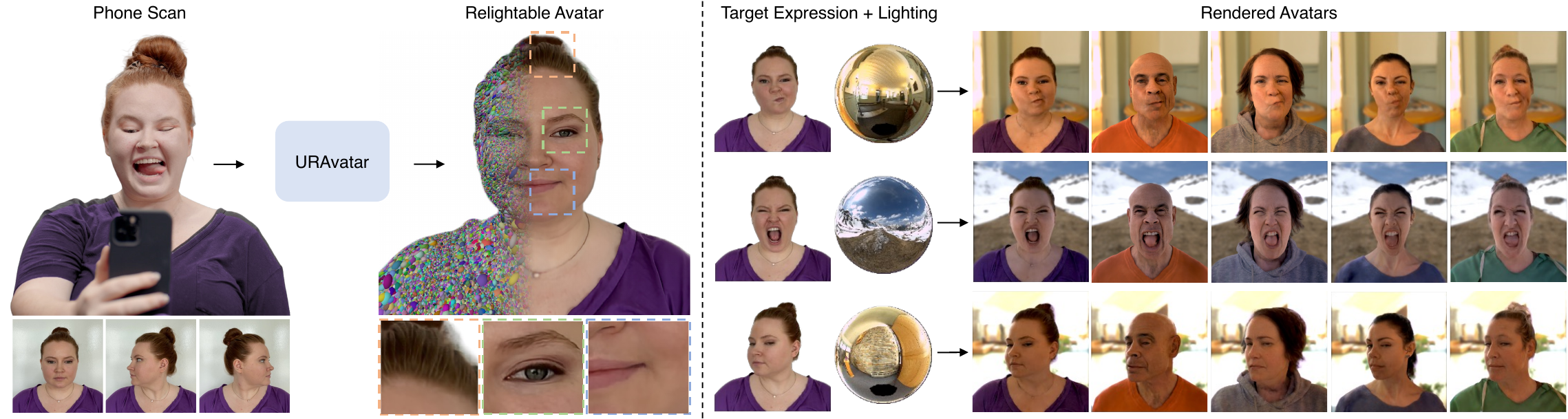}
  \caption{URAvatar. Our approach enables the creation of drivable and relightable photorealistic head avatars from a single phone scan (left). The reconstructed avatars can be driven consistently across identities under different illuminations in real time (right). \url{https://junxuan-li.github.io/urgca-website/}}
  \label{fig:teaser}
\end{teaserfigure}


\maketitle

\section{Introduction}

Photorealistic head avatars are fundamental to enabling communication in virtual environments~\cite{lombardi2018deep,ma2021pixel}. To establish coherent presence in such environments, avatars have to be illuminated to match the particular environment that they are in. Consider a virtual interaction in a room with natural light shining in from a side window. If the avatars in the scene are uniformly lit, or lit as if they were in a room with ceiling fluorescent light, the incongruence between environment and avatars will interfere with--and likely break--the sense of presence.

\begin{figure*}
    \centering
    \includegraphics[width=\textwidth] {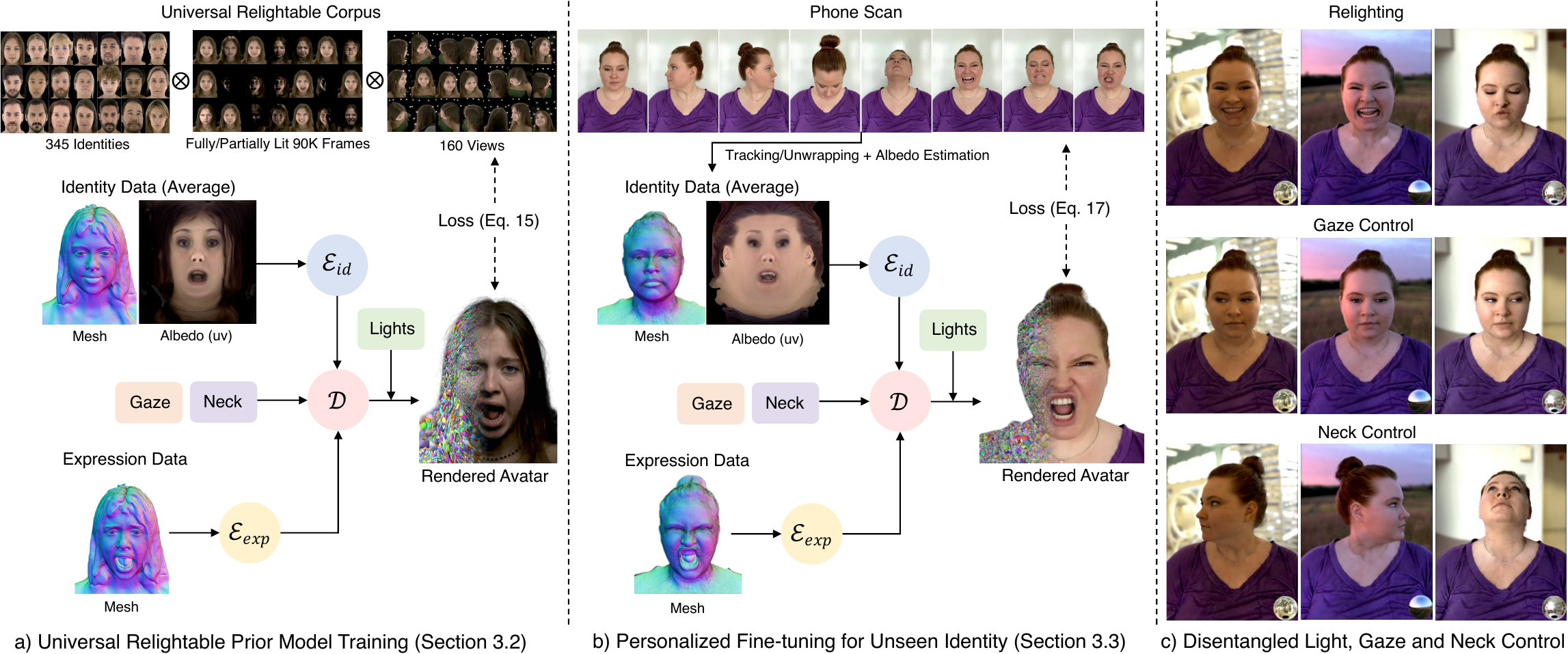}
    \caption{Method Overview.
    (a) We employ a large relightable corpus of multi-view facial performances to train a cross-identity decoder $\mathcal{D}$ that can generate volumetric avatar representations.
    (b) Given a single phone scan of an unseen identity, we reconstruct the head pose, geometry, and albedo texture, and fine-tune our pretrained relightable prior model.
    (c) Our final model provides disentangled control over relighting, gaze and neck control.
    }
    \label{fig:overview}
\end{figure*}

The challenge is that human heads are among the most complex objects to relight accurately. Light interacts with the head in varied ways, scattering in the skin, reflecting in the eyes and teeth, getting trapped in hair strands, and so on. This complexity is compounded by the diversity among human beings in facial structure, skin types, eye colors, accessories, and hair types. Traditionally, measuring scattering and reflectance properties to build authentic relightable avatars has required detailed scans in multi-light capture systems. \cite{debevec2000acquiring,ghosh2011multiview,bi2021deep,saito2024rgca} 
Such capture systems are costly and require specialists to build. The scans themselves are time-consuming and inconvenient. To truly build virtual communities that the majority of people can access, we require the means to quickly and effortlessly create relightable avatars, across the span of human diversity.  

Recent approaches have attempted to drastically reduce the capture data to as little as a single input image \cite{yamaguchi2018high,lattas2020avatarme} or a monocular video \cite{wang2023sunstage,bharadwaj2023flare}. Yet, there remains a clear fidelity gap between the studio-captured avatars and the ones from lightweight inputs.
In this paper, our goal is to achieve comparable relightable quality to those studio-captured avatars from just a single phone scan.

To close the quality gap, we introduce \emph{URAvatar} (pronounced ``your avatar''), a \textbf{U}niversal \textbf{R}elightable \textbf{Avatar} prior learned from hundreds of individuals captured with a multi-view and multi-light capture system in an end-to-end manner.
URAvatar uses a set of 3D Gaussians \cite{kerbl20233d} to represent the intricate geometry of human heads and hair, and builds a prior on the joint distribution of identity, expressions, and illumination.
This enables the modeling of a relightable and drivable avatar with high-fidelity details from under-constrained input as shown in \cref{fig:teaser}.
Unlike existing approaches that learn priors based on parametric BDRFs \cite{yamaguchi2018high,smith2020morphable,Li2020LearningFormation,lattas2021avatarme}, we build our relightable appearance prior based on learnable radiance transfer \cite{saito2024rgca} that incorporates global light transport as a result of multi-bounce scattering and reflection.
This way, we can efficiently relight avatars with global illumination under various environments without expensive ray tracing.
Moreover, the model can be directly supervised to reproduce the ground-truth images without being restricted by the expressiveness of the chosen BRDF model.
For consistent drivability across identities, we balance between the explicitness of control and the scalability of training.
In particular, we choose to explicitly model eye gaze and neck rotation in the form of linear blend skinning, as they can be reliably tracked.
On the other hand, facial expressions, including complex tongue motions, are all learned as latent codes in a self-supervised manner \cite{lombardi2018deep,xu2023latentavatar}. 

Once trained, we finetune the avatar with an input phone scan of a new person.
We reduce the domain gap between the pretrained model and the phone scan by estimating the albedo in screen space and unwrapping it to UV space for identity conditioning.
Then, we estimate illumination by regression and refine it via inverse rendering.
Finally, the weights of the prior model are updated to best explain the phone scan via inverse rendering.
Our carefully designed finetuning strategy ensures that the relightability is retained from the prior, while recovering essential person-specific details.

To measure the fidelity of our approach, we collect ground-truth relighting data under various continuous illumination conditions with a capture dome that consists of multiple LED screens.
This allows us to quantitatively compare the synthesis and real-world observations given known natural illumination.
Our experiments show that our approach outperforms prior methods by a large margin, and clearly demonstrates the efficacy of our prior-based relighting that accounts for global light transport in real time.

Our principal contributions are:
\begin{enumerate}[topsep=0pt]
    \item We introduce a universal relightable avatar prior model learned from hundreds of dynamic performance captures with a multi-view and multi-light system.
    \item We build a drivable head avatar from a phone scan that can be rendered and relit with global light transport in real-time.
    \item A capture system and evaluation protocol to measure the accuracy of relighting under continuous illuminations.
\end{enumerate}

\section{Related Work}

Authentic avatars for mass adoption must satisfy the following criteria: they must be drivable, relightable, and lightweight enough for anyone to create.
In what follows, we discuss prior works based on these criteria.

\subsection{Drivable Avatars}
In computer graphics, controlling facial expressions of avatars has been primarily driven by visual effects and games.
To enable consistent control across identities, anatomically motivated FACS action units \cite{friesen1978facial} are widely used as the basis of blendshapes \cite{lewis2014practice}.
However, this basis is often insufficient to capture person-specific variations, and tends to require additional correctives \cite{li2013realtime}.
Data-driven approaches construct linear \cite{pighin2006synthesizing,blanz2023morphable}, multi-linear \cite{vlasic2006face,cao2013facewarehouse}, and non-linear \cite{ranjan2018generating,tran2018nonlinear} bases from captured 3D data.
The FLAME model \cite{li2017learning} also incorporates linear blend skinning (LBS) for jaw and neck motions.
These approaches lack fine-grained subtle expressions as well as tongue and eye motions.
Deep appearance models \cite{lombardi2018deep} propose a self-supervised method to discover the expression latent space using variational autoencoders (VAEs).
This approach allows the driving of authentic facial expressions of users in a fully data-driven manner.
Later, LatentAvatar \cite{xu2023latentavatar} shows that a similar construction is possible for less constrained setups.
\citet{cao2022authentic} learn the latent expression space across multiple identities to enable semantically consistent driving while retaining person-specific expressions.
While this approach works well for relatively small deformations, large deformations caused by articulations, such as neck or eye motions, lead to undesired artifacts.
To address this, our work combines the latent expression codes with explicit eye models \cite{schwartz2020eyes,li2022eyenerf} and neck articulations via LBS \cite{li2017learning}, further enhancing the drivability and fidelity.

\subsection{Lightweight Avatar Generation}

Early works on photorealistic human digitization required dedicated reconstruction pipelines and capture systems for individual components including hair \cite{luo2013structure,echevarria2014capturing,nam2019strand}, face\cite{debevec2000acquiring,ghosh2011multiview}, eyes \cite{berard2014high}, and inner mouth \cite{wu2016model}.
While these approaches are non-trivial to scale for large-scale identities, \citet{ichim2015dynamic} show the promise of reconstructing personalized avatars from a phone scan.
Follow-up works support more diverse hair styles \cite{cao2016real} or enable reconstruction from a single image \cite{hu2017avatar,nagano2018pagan}.
However, these approaches tend to lack photorealism due to the limited expressiveness of the underlying morphable models and/or mesh representations.
More recently, neural fields \cite{xie2022neural}, including NeRF~\cite{mildenhall2021nerf}, show remarkable progress on modeling complex geometry and appearance.
This is also extended to avatar reconstructions from casually captured video data \cite{Zheng2021IMA,gafni2021dynamic,grassal2022neural,zielonka2023instant,athar2022rignerf,athar2023flame}.
While NeRF and its variants are slow to render, neural rendering approaches based on Mixture of Volumetric Primitives \cite{lombardi2021mixture,cao2022authentic} or Neural Deferred Rendering \cite{thies2019deferred,wang2023styleavatar} show the ability to render 3D avatars in real-time for interactive applications.
Despite impressive progress on modeling authentic avatars from lightweight inputs, the common limitation of these approaches is that the illumination is baked into the appearance model and avatars cannot be relit under different environments.

\subsection{Avatar Relighting}

Relighting is a critical property to enable photorealistic composition of avatars into a scene.
\citet{debevec2000acquiring} show that one-light-at-a-time (OLAT) capture can be used to recover reflectance fields.
The follow-up work further supports dynamics relighting \cite{peers2007post} and accelerates the acquisition process by leveraging spherical gradient illuminations \cite{ma2007rapid,ghosh2011multiview,fyffe2016near}.
Despite high-fidelity outputs, it remains non-trivial to widely adopt such a system.
While the work by \citet{sengupta2021light} reduces the hardware requirement to a single camera and a single monitor, it supports neither reanimation nor novel-view synthesis.
While portrait relighting approaches \cite{sun2019single,tewari2020stylerig,wang2020single,pandey2021total,yeh2022learning,zhang2021neural,kim2024switchlight,meka2019deep,meka2020deep} support relighting in screen space, due to the lack of temporal or 3D information, they tend to produce flickering when changing view or expressions.
Some approaches \cite{tan2022volux,mei2024holo} use 3D-aware GANs to synthesize relightable faces, but they have limited animatability.
Another common approach for relighting is to estimate skin reflectance properties such as albedo, roughness, and surface normals from multi-view images \cite{Li2020LearningFormation,liu2022rapid} or a single image \cite{yamaguchi2018high,chen2019photo,lattas2020avatarme,lin2023single}.
Once estimated, the avatars can be relit with path tracing or real-time shaders.
However, these approaches are limited to skin regions and non-trivial to unify for the entire head due to complex scattering/reflectance properties of different components including hair, eyes, teeth, and skin.
Incorporating intrinsic decomposition into the image formation process of GANs also achieves relighting in the wild \cite{ranjan2023facelit,deng2023lumigan}.
Optimization-based approaches from a phone scan demonstrate personalized relightable avatar reconstruction \cite{wang2023sunstage,zheng2023pointavatar,bharadwaj2023flare}.
While they show promising results, fine-grained driving remains challenging and photorealism under novel illumination is still limited due to the limited prior knowledge about light transport. 
Recent neural relighting approaches show remarkable progress in terms of photorealism \cite{bi2021deep,yang2023towards,saito2024rgca,xu2024artist}.
MEGANE \cite{li2023megane} learns a relightable appearance model across multiple identities.
URHand \cite{chen2024urhand} enables the instant personalization of a pretrained relightable hand prior.
VRMM \cite{yang2024vrmm} concurrently proposes a multi-identity relightable avatar model based on MVP~\cite{lombardi2021mixture} and a linear lighting model~\cite{yang2023towards}. However, it remains a challenge to faithfully capture geometric details as hair strands using mesh \cite{chen2024urhand} or Mixture of Volumetric Primitives \cite{lombardi2021mixture,li2023megane,yang2024vrmm}.
In this work, we base our geometric and appearance representation on 3D Gaussians \cite{kerbl20233d} and learnable radiance transfer \cite{sloan2002precomputed,saito2024rgca}, respectively.
Our approach enables, for the first time, the learning of a universal relightable prior that natively supports real-time relighting with global light transport under various illumination.  
In addition, our approach enables personalization from a phone capture with unknown illumination.

\begin{figure*}
\centering
\includegraphics[width=0.98\linewidth]{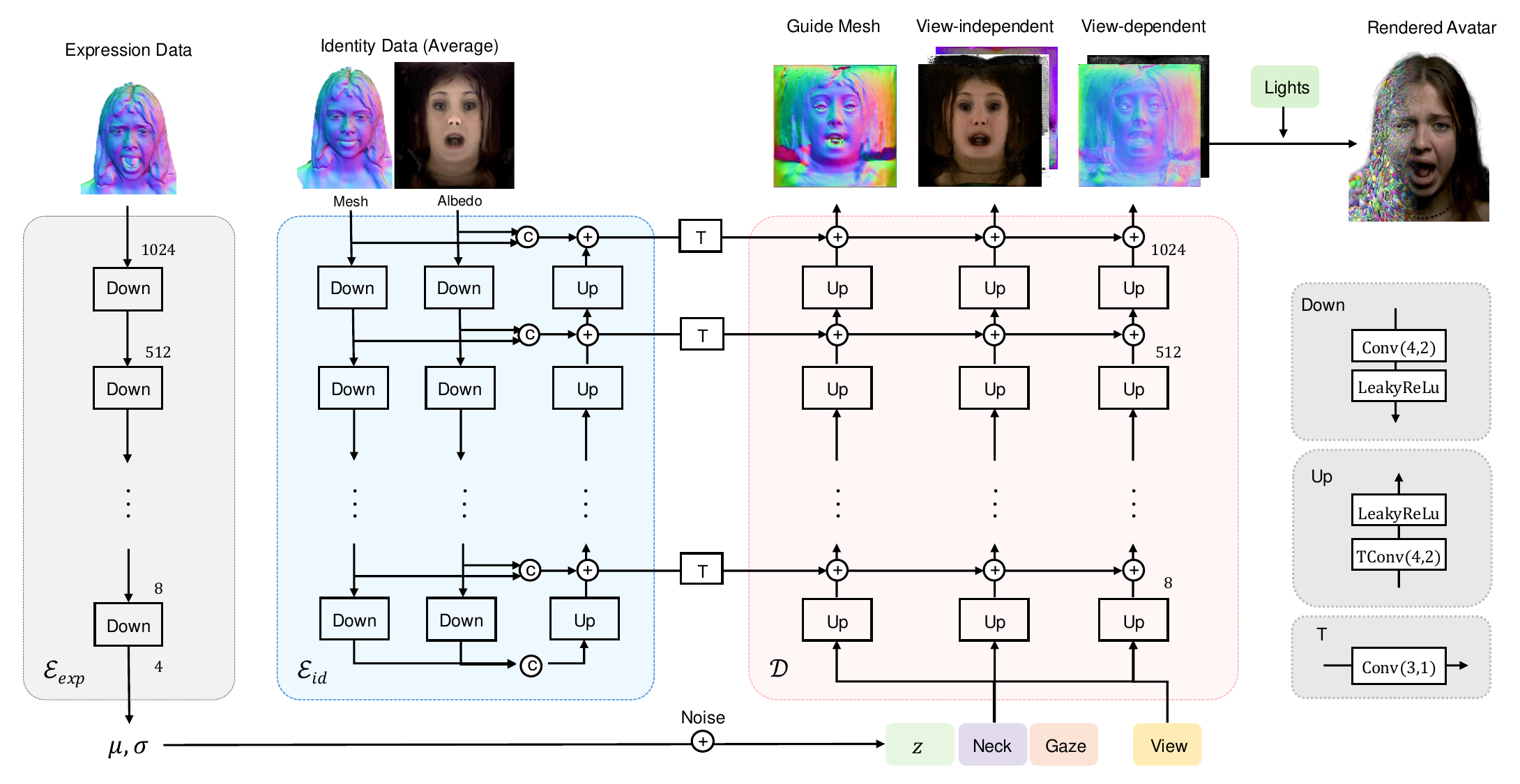}
\caption{Network architecture. Our expression encoder, $\gE_{\text{exp}}$, takes a 1024x1024 positional map of face geometry as input and encodes it into an expression latent code with the map size of 4x4. Our downsampling block consists of a convolutional layer with a kernel size of 4 and stride of 2, followed by a leaky ReLU activation function.  Similarly, our upsampling block is composed of a transposed convolutional layer with a kernel size of 4 and stride of 2, followed by a leaky ReLU activation function. Our identity encoder, $\gE_{\text{id}}$, is a U-Net-like architecture that takes the mean texture and geometry of a subject as input, producing a multi-scale feature pyramid as the ID conditioning data. The produced feature maps are then added to the corresponding layer of the decoder to produce the guide mesh and its Gaussian parameters. Our decoder consists of 7 upsampling blocks that take a map with a size of 4x4 as input and output 1024x1024 Gaussian parameter maps.}
\label{fig:network_arch}
\end{figure*}

\section{Method}

As our geometry and appearance representations are based on Relightable Gaussian Codec Avatars \cite{saito2024rgca}, we first describe the foundation of 3D Gaussians and learnable radiance transfer.
We then discuss how we extend Gaussian Codec Avatars to build a universal relightable prior with multi-identity training data.
Finally, we provide the details of our finetuning approach to create a personalized relightable model from a phone scan using the universal relightable prior.

\subsection{Preliminaries: Relightable 3D Gaussians}
\label{sect:gaussians}
Avatars are represented as a collection of 3D Gaussians, denoted as $\mathbf{g}_k=\{\mathbf{t}_k, \mathbf{q}_k, \mathbf{s}_k, o_k, \mathbf{c}_k\}$.
The parameters include a translation vector $\mathbf{t}_k \in \mathbb{R}^3$, rotation parameterized by a unit quaternion $\mathbf{q}_k \in \mathbb{R}^4$, scale factors $\mathbf{s}_k \in \mathbb{R}_+^3$ along three orthogonal axis, an opacity value $o_k \in \mathbb{R}_+$, and a color $\mathbf{c}_k \in \mathbb{R}_+^3$.
Following 3D Gaussian Splatting \cite{kerbl20233d}, the Gaussians can be efficiently rendered at high resolution in real-time.

We base our appearance model on learnable radiance transfer \cite{saito2024rgca}.
To model the appearance under different illuminations, precomputed radiance transfer (PRT) \cite{sloan2002precomputed,wang2009all} decomposes the integral of the rendering equation into the product of extrinsic illumination $L(\boldsymbol{\omega})$ and intrinsic radiance transfer $T(\mathbf{p}, \boldsymbol{\omega}, \boldsymbol{\omega}^o)$.
\citeauthor{saito2024rgca} further extend PRT by directly learning the parameters of the transfer function from multi-view and multi-light capture data,
and decomposing $T(\mathbf{p}, \boldsymbol{\omega}, \boldsymbol{\omega}^o)$ into diffuse terms (independent of the viewing direction) and specular terms:
\begin{align}
    \begin{aligned}
\mathbf{c}(\mathbf{p}, \boldsymbol{\omega}^o) &=\int_{\mathbb{S}^2} L(\boldsymbol{\omega}) \cdot T(\mathbf{p}, \boldsymbol{\omega},\boldsymbol{\omega}^o) d \boldsymbol{\omega}, \\
&= \int_{\mathbb{S}^2} L(\boldsymbol{\omega}) \cdot \left(T^\textrm{diffuse}(\mathbf{p}, \boldsymbol{\omega}) +  T^\textrm{specular}(\mathbf{p}, \boldsymbol{\omega}, \boldsymbol{\omega}^o) \right) d \boldsymbol{\omega} \text{,}
\end{aligned}
\end{align}
where $\mathbf{c}$ is the outgoing radiance at the position $p$ along $\boldsymbol{\omega}^o$, and $L$ incoming light. 
In particular, the outgoing radiance for each Gaussian $\mathbf{c}_k$ is decomposed into view-independent diffuse and view-dependent specular terms, represented as $\mathbf{c}_k = \mathbf{c}^\text{diffuse}_{k} + \mathbf{c}^\text{specular}_{k}$.
The diffuse color is calculated through the integration of the incoming radiance and the intrinsic radiance transfer, both of which are parameterized by spherical harmonics (SH) \cite{sloan2002precomputed}:
\begin{align}
\label{eq:diff}
\mathbf{c}^\text{diffuse}_{k} = \boldsymbol{\rho}_k\odot\sum_{i=1}^{(n+1)^2}{\mathbf{L}_{i}\odot\mathbf{d}_{k}^{i}} \text{,}
\end{align}
where $\mathbf{L}_{i}$ denotes the $i$-th element in $n$-th order spherical harmonics (SH) coefficients of the incident lights, $\mathbf{d}^{i}_{k}$ represents the $i$-th element in $n$-th order SH coefficients of the learnable radiance transfer function, and $\boldsymbol{\rho}_k$ is the base albedo color.
These terms are modeled individually for RGB channels.
Inspired by \citet{wang2009all}, the specular reflection is represented as spherical Gaussians $G_s(\boldsymbol{\omega}; \mathbf{a}, \sigma)$ with the central direction of the lobe $\mathbf{a}$ and roughness $\sigma$:
\begin{align}
\label{eq:spec}
\mathbf{c}^{\textrm{specular}}_{k}(\boldsymbol{\omega}^o_k) &= v_k(\boldsymbol{\omega}^o_k) \int_{\mathbb{S}^2} \mathbf{L}(\boldsymbol{\omega})G_s(\boldsymbol{\omega}; \mathbf{a}_k, \sigma_{k}) \mathrm{d} \boldsymbol{\omega}, \\
\mathbf{a}_k &= 2(\boldsymbol{\omega}^o_k \cdot \mathbf{n}_k)\mathbf{n}_k - \boldsymbol{\omega}^o_k \text{.}
\end{align}
Here, $v_k(\boldsymbol{\omega}^o_k) \in (0,1)$ is a learnable view-dependent visibility term that accounts for Fresnel reflection, occlusion, and geometric attenuation, $\boldsymbol{\omega}^o_k\in\mathbb{R}^3$ is the viewing direction evaluated at the Gaussian center, and $\mathbf{n}_k$ is a view-dependent normal for each Gaussian.

\subsection{Universal Relightable Prior Model}

Inspired by prior work \cite{cao2022authentic}, we employ an identity-conditioned hypernetwork \cite{ha2016hypernetworks} to generate person-specific avatars.
In particular, the hypernetwork takes identity features as input, and produces a subset of person-specific network weights for each subject's avatar decoder.
This decoder produces relightable 3D Gaussians 
corresponding to the input head state (facial expression, gaze direction, and neck rotation), and input lighting environment and viewpoint.
We show the overview in \cref{fig:overview} (a).

\subsubsection{Identity-conditioned Hypernetwork}
To allow extraction of high-frequency person-specific details, our identity encoder $\gE_{\text{id}}$ takes identity features in the form of a mean albedo texture map $\mT_{\text{mean}}$ and a mean geometry map $\mG_{\text{mean}}$ unwrapped in a $1024^2$ UV space as input, and produces `untied' bias maps $\Theta^{\text{id}}_{\text{g}}, \Theta^{\text{id}}_{\text{fi}}, \Theta^{\text{id}}_{\text{fv}}$. These bias maps are injected at various levels of the decoding architecture, described below. Our hypernetwork also produces an expression-agnostic opacity $\{ o_k\}_{k=1}^{M}$ and albedo $\{ \boldsymbol{\rho}_k\}_{k=1}^{M}$ for the 3D Gaussians. Formally, the identity encoder is defined as:
\begin{align}
\label{eq:idencoder}
\Theta^{\text{id}}_{\text{g}}, \Theta^{\text{id}}_{\text{fi}}, \Theta^{\text{id}}_{\text{fv}}, \{ o_k, \boldsymbol{\rho}_k\}_{k=1}^{M} = \gE_{\text{id}} (\mT_{\text{mean}}, \mG_{\text{mean}}; \Phi_{\text{id}} ) \text{.}
\end{align}

\subsubsection{Expression Encoder}
We use a variational autoencoder to model a shared latent distribution of facial expressions across identities.
To avoid the domain shift between studio and phone-captured textures, our expression encoder $\gE_{\text{exp}}$ takes only the difference of geometry maps as input, i.e., 
$\Delta \mG_{\text{exp}} = \mG_{\text{exp}} - \mG_{\text{mean}}$, where $\mG_\text{exp}$ is the current and $\mG_\text{mean}$ is the mean geometry.
To preserve subtle facial expression details when using geometry-only inputs, we use high-quality tracking (\cref{sec:hrgeo}) to generate these maps.
We then generate a universal expression latent code $\rvz \in \R^{256}$ as follows:
\begin{align}
\boldsymbol{\mu},\boldsymbol{\sigma} &= \gE_{\text{exp}} (\Delta \mG_{\text{exp}} ; \Phi_{\text{exp}}) \text{,} \\
\rvz &= \boldsymbol{\mu} + \boldsymbol{\sigma} \cdot \gN(0,1) \text{,}
\end{align}
where $\gN(0,1)$ is the unit normal distribution.
Since the expression latent code is trained in an end-to-end manner with multi-identity data, once the model is trained, the same expression code can be applied to different identities for driving.

\subsubsection{Avatar Decoder}
Building upon the foundations laid by previous work \cite{saito2024rgca,cao2022authentic,lombardi2021mixture,li2017learning}, we parameterize and anchor Gaussians on a guide mesh and model the facial expressions in a canonical space. We further expand the geometry to encompass the shoulder region and use predefined linear blend skinning to model neck rotations.
A geometry decoder $\gD_{\text{g}}$ produces the vertices $\{\hat{\rvt}_k\}_{k=1}^{M}$ of this extended guide mesh:
\begin{align}
\label{eq:decoderg}
\{\hat{\rvt}_k\}_{k=1}^{M} = \mathcal{D}_{\text{g}}(\mathbf{z}, \mathbf{e}_{\{l,r\}}, \mathbf{r}_{\text{n}}; \Theta^{\text{id}}_{\text{g}}, \Phi_{\text{g}}) \text{,}
\end{align}
where $\rvz$ is the expression code, $\mathbf{e}_{\{l,r\}} \in \R^3$ are eye gaze direction vectors, $\rvr_{\text{n}} \in \R^3$ denotes the axis-angle neck rotation relative to the head, and
$ \Theta^{\text{id}}_{\text{g}} $ is an identity-specific bias from \cref{eq:idencoder}.

We split the appearance model into view-independent and view-dependent components.
Our view-independent face relightable Gaussian decoder, $\gD_{\text{fi}}$, takes expression code, gaze vectors, and neck rotation as input.
It is further conditioned with the identity `untied' bias map $\Theta^{\text{id}}_{\text{fi}}$ derived from the identity encoder, resulting in the output of view-independent attributes for each 3D Gaussian:
\begin{align}
\label{eq:decoderci}
\{\mathbf{\delta t}_k, \mathbf{q}_k, \mathbf{s}_k, \mathbf{d}^{\textrm{c}}_{k}, \mathbf{d}^{\textrm{m}}_{k}, \sigma_{k}\}_{k=1}^{M} = \mathcal{D}_{\text{fi}}(\mathbf{z}, \mathbf{e}_{\{l,r\}}, \mathbf{r}_{\text{n}}; \Theta^{\text{id}}_{\text{fi}}, \Phi_{\text{fi}}) \text{.}
\end{align}
Here, $\mathbf{d}^{\textrm{c}}_{k}$ and $\mathbf{d}^{\textrm{m}}_{k}$ are color and monochrome SH coefficients \cite{saito2024rgca}, respectively, and $\sigma_k$ is the roughness parameter defined in \cref{eq:spec}.
Our view-dependent face relightable Gaussian decoders, $\gD_{\text{fv}}$, incorporates the view direction to the head center, $\boldsymbol{\omega}_o$, as additional input, subsequently generating view-dependent delta normal and visibility terms for each Gaussian:
\begin{align}
\label{eq:decodercv}
\{\mathbf{\delta n}_k, v_k\}_{k=1}^{M} = \mathcal{D}_{\text{fv}}(\mathbf{z}, \mathbf{e}_{\{l,r\}}, \mathbf{r}_{\text{n}}, \boldsymbol{\omega}_o; \Theta^{\text{id}}_{\text{fv}}, \Phi_{\text{fv}}).
\end{align}
Here, $\Phi_{\text{fi}}$ and $\Phi_{\text{fv}}$ represent the learnable parameters of each respective decoder.
The final Gaussian position is a composite of the guide mesh's vertex positions, as derived from \cref{eq:decoderg}, and the delta position from \cref{eq:decoderci}, taking the form $\rvt_k = \hat{\rvt}_k + \mathbf{\delta t}_k$.
The view-dependent surface normal of each Gaussian, $\rvn_k$, is a composition of the mesh normal, derived from the guided mesh, and the delta normal decoded from \cref{eq:decodercv}, taking the form $\rvn_k = \hat{\rvn}_k + \mathbf{\delta n}_k$.

\subsubsection{Universal Relightable Explicit Eye Model}
We adapt previous works \cite{saito2024rgca,schwartz2020eyes} for modeling the eye as a universal relightable explicit eye model.
We use the same network architecture as the encoder $\gE_{\text{id}}$ for each identity's eye encoder $\gE_{\text{eye}}$, defined as below:
\begin{align}
\Theta^{\text{eye}}_{\text{ei}}, \Theta^{\text{eye}}_{\text{ea}} = \gE_{\text{eye}} (\mT_{\text{eye}}, \mG_{\text{eye}}; \Phi_{\text{eye}} ) \text{,}
\end{align}
where $\mT_{\text{eye}}$ and $\mG_{\text{eye}}$ are the cropped eye regions from the mean texture and mean geometry maps $\mT_{\text{mean}}$ and $\mG_{\text{mean}}$.
The output of this encoder is the `untied' bias map of each level of the eye's decoder.
\looseness-1

During the phone capture fine-tuning stage, we observed that the networks were unable to preserve the prior knowledge acquired during the pre-training stage and failed to reproduce eye glint after fine-tuning, due to the limited observation of the eye regions from phone captures.
Consequently, we propose a unified specular visibility decoder $\gD_{\text{ev}}$, which does not require any conditional information as input.
This design is intended to encourage the network to learn an eye reflection model that can easily generalize to unseen identities.
Therefore, our universal relightable eye decoder is defined as follows:
\begin{align}
\{\mathbf{q}_k, \mathbf{s}_k, o_k,\mathbf{d}^{\textrm{c}}_{k}, \mathbf{d}^{\textrm{m}}_{k}, \sigma_{k}\}_{k=1}^{M_e} &= \mathcal{D}_{\text{ei}}( \mathbf{e}_{\{l,r\}}; \Theta^{\text{eye}}_{\text{ei}}, \Phi_{\text{ei}}) \text{,} \\
\{\boldsymbol{\rho}_k\}_{k=1}^{M_e} &= \mathcal{D}_{\text{ea}}(\mathbf{e}_{\{l,r\}}, \boldsymbol{\omega}_o; \Theta^{\text{eye}}_{\text{ea}}, \Phi_{\text{ea}}) \text{,}\\
\{v_k\}_{k=1}^{M_e} &= \mathcal{D}_{\text{ev}}(\mathbf{e}_{\{l,r\}}, \boldsymbol{\omega}_o; \Phi_{\text{ev}}) \text{.}
\end{align}
\subsubsection{High-Quality Tracking Geometry}\label{sec:hrgeo}
We first track each subject independently using a high-quality template head mesh.
This mesh is subsequently used to supervise our geometry decoder's output $\hat{\rvt}_k$.
We observed that this form of geometry supervision plays a crucial role in preventing 3D Gaussians from getting trapped in local minima during the early stages of universal relightable prior model training.
To obtain this tracked mesh, we modify the Pixel Codec Avatar (PiCA) \cite{ma2021pixel} architecture to extend coverage to the upper body, specifically the head, neck, and shoulders.
For each identity in our training dataset, we therefore fit a personalized pixel codec avatar using inverse rendering to reconstruct all the fully-lit, multi-view input images.
Subsequently, we extract the geometry for each frame from PiCA's geometry branch.
Leveraging the dense position map-based geometry representation and per-pixel color decoding, we can reconstruct high-quality geometry for each frame.
For monocular phone scans, we follow a pipeline similar to \citet{cao2022authentic}, but replace the coarse base geometry with the aforementioned high-resolution template mesh and photometric refinement \cite{ma2021pixel}.

\subsubsection{Conditioning Albedo Texture Acquisition}\label{sec:albedotex}

We use an off-the-shelf portrait relighting method \cite{kim2024switchlight} to estimate the illumination and the albedo of input images of the face.
The estimated illumination is used as an initialization for the environment light fitting in \cref{sec:light}.
To maximize dataset consistency, we use the same algorithm to extract albedo in both studio and phone settings.
These albedo images are then unwrapped onto the high-quality tracked mesh to obtain the mean albedo texture $\mT_\text{mean}$.
This albedo texture is used as the conditioning input to the identity encoder in \cref{eq:idencoder}.
During experiments, we found that utilizing this mean albedo texture, in contrast to a simple color transformation \cite{cao2022authentic}, assists in more effectively de-lighting the face under varying illumination conditions and offers a more consistent conditional input for our identity encoder.

\subsubsection{Training Losses}
Given multi-view video data of a person illuminated with known point light patterns, we follow previous works \cite{saito2024rgca,kerbl20233d} to use the same loss functions to optimize all trainable network parameters $\Phi$ with the following loss function:
\begin{align}
\mathcal{L} = \mathcal{L}_{\mathrm{rec}} + \mathcal{L}_{\mathrm{reg}} + \lambda_{\mathrm{kl}} \mathcal{L}_{\mathrm{kl}} \text{,}
\end{align}
where
$\mathcal{L}_{\mathrm{rec}}$ is the reconstruction loss consists of L1 and SSIM on rendered images and L2 loss on the high quality tracking geometry, $\mathcal{L}_{\mathrm{reg}}$ is the set of loss functions regularizing the scale of Gaussians, negative color values, and scale, opacity, and visibility of eye Gaussians.
$\mathcal{L}_{\mathrm{kl}}$ is the KL-divergence loss on expression code $\rvz$. 
Except the geometry loss, where we use the relative weight of $0.1$, we use the identical relative weights to \citet{saito2024rgca}. Please refer to \citet{saito2024rgca} for the details of each loss function.

\begin{figure}
  \centering
  \includegraphics[width=\linewidth]{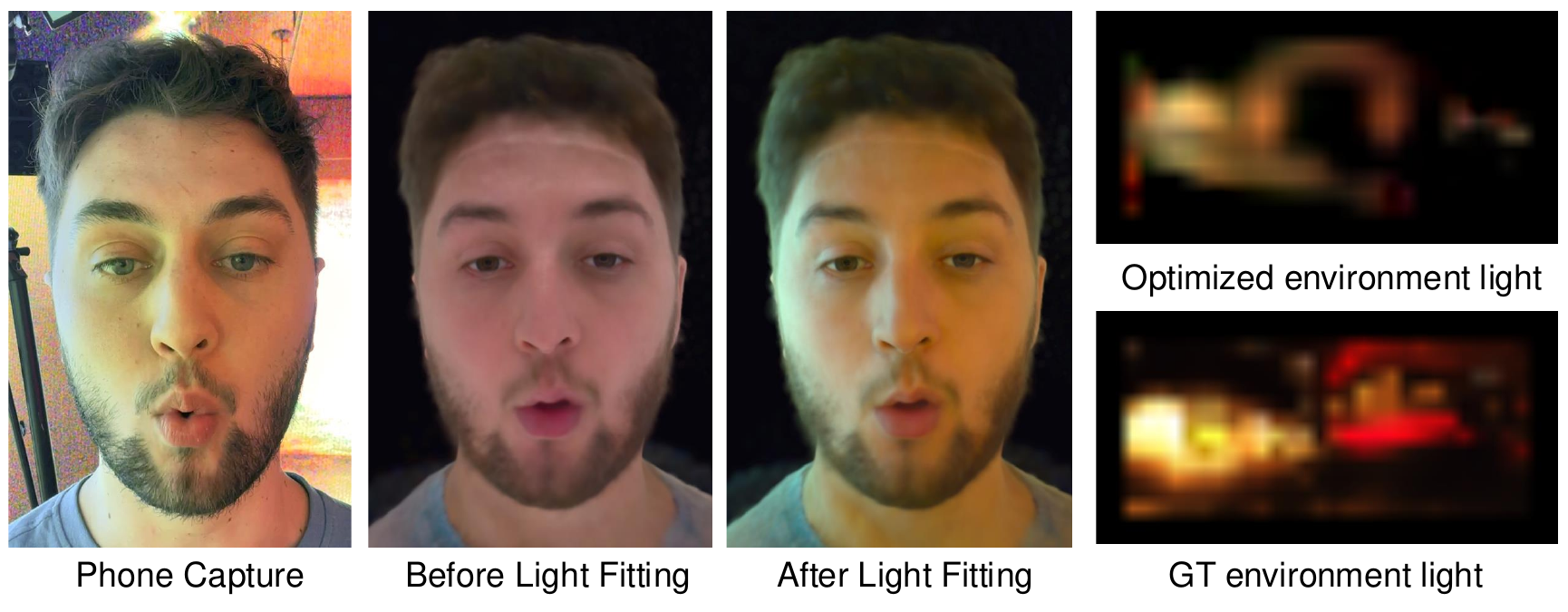}
  \caption{Visualization of the effect of our fitted environment lights, and the comparison to the ground-truth environment lights. }
  \label{fig:fit-light}
\end{figure}

\begin{figure*}
     \centering
     \begin{subfigure}[b]{0.49\textwidth}
         \centering
         \includegraphics[width=\textwidth]{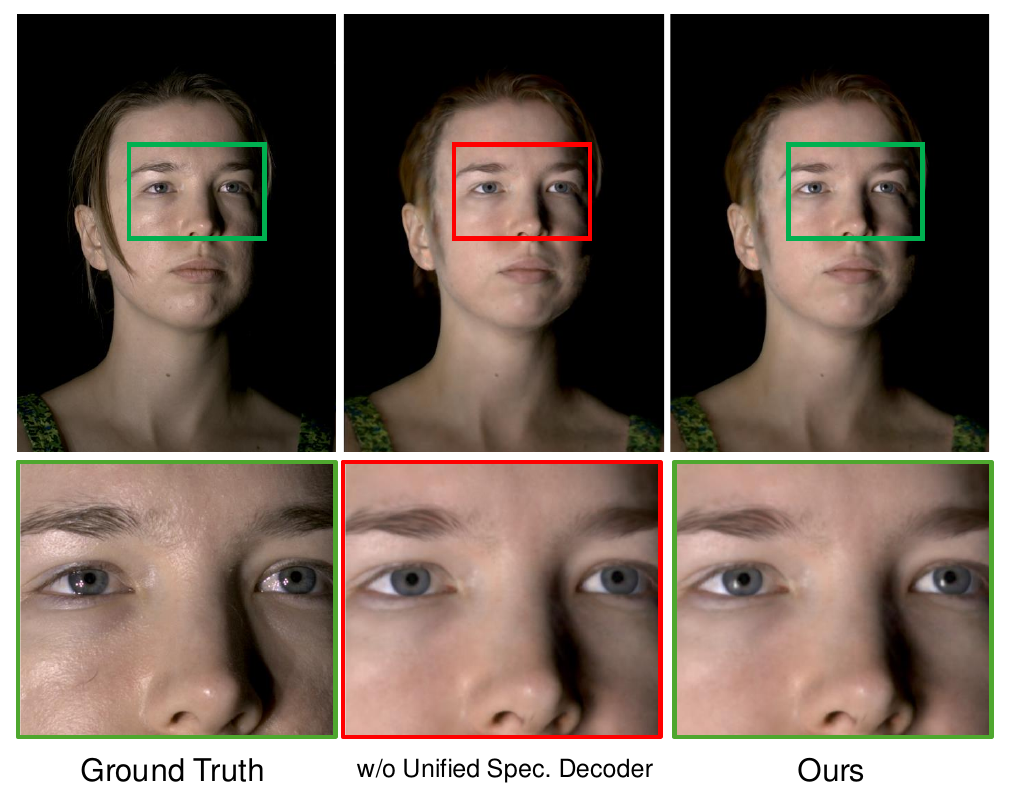}
         \caption{Comparison on w vs. w/o using unified specular visibility decoder for eyes. }
         \label{fig:eye_glint_compare}
     \end{subfigure}
     \hfill
     \begin{subfigure}[b]{0.49\textwidth}
         \centering
         \includegraphics[width=\textwidth]{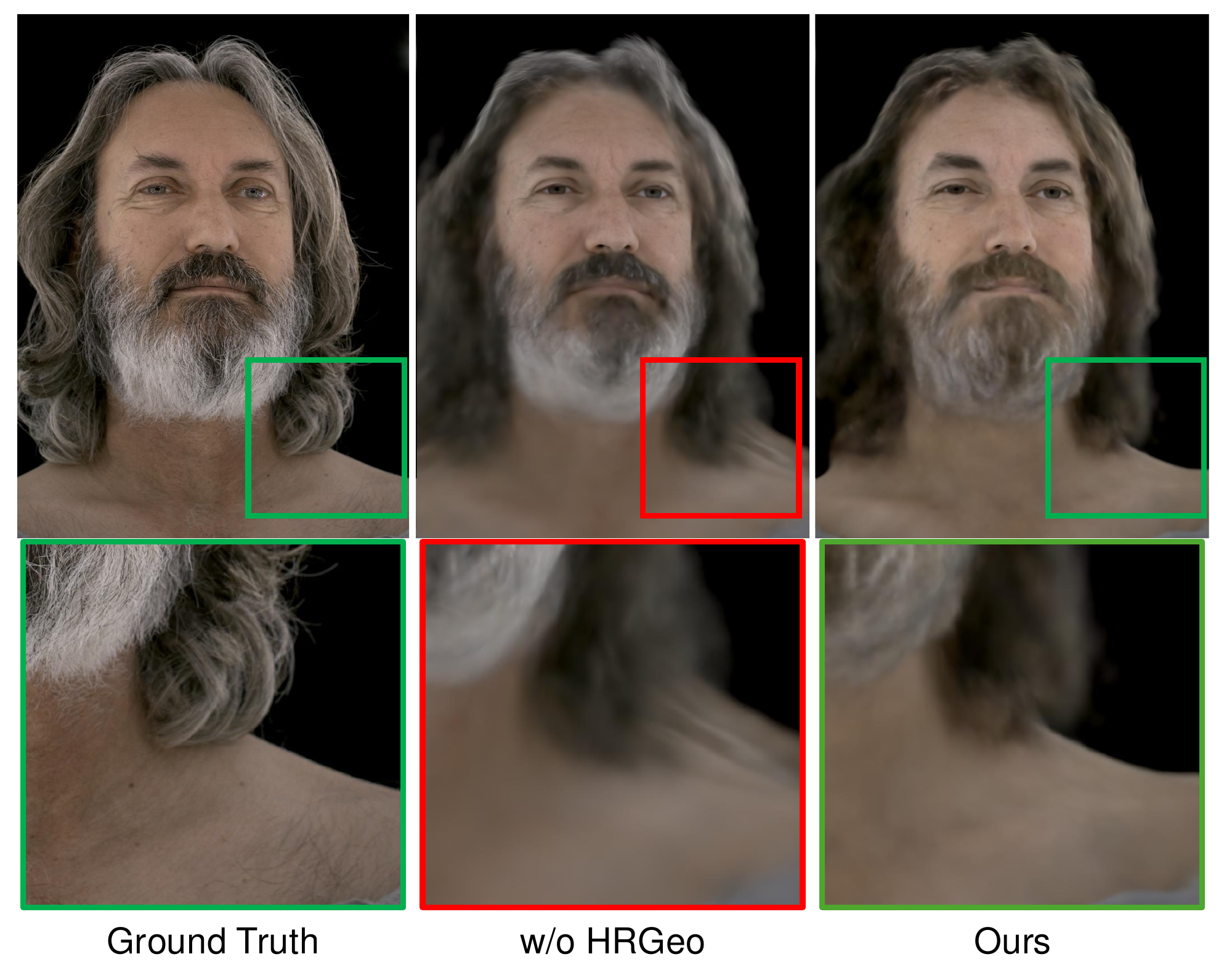}
         \caption{Comparison on w vs. w/o using high-resolution tracked mesh (HRGeo).  }
         \label{fig:pica_compare}
     \end{subfigure}
     \caption{Ablation study on unified eye specular visibility decoder and high-resolution tracked mesh.}
\label{fig:Ablation}
\end{figure*}
\subsection{Personalized Avatar from Phone Scan}
\label{sec:personalized_avatar}
With the pre-trained universal relightable prior model, we further fine-tune it using phone capture data to generate a personalized relightable avatar.
During the fine-tuning stage, we initially freeze the pre-trained encoders and decoder, optimizing the environment lights to achieve a low reconstruction loss on the face.
Subsequently, we freeze the environment lights and fine-tune the encoders and decoders for a few iterations to improve the likeness of the captured subject.
See \cref{fig:overview} (b) for an overview.
For preprocessing the phone capture data, we apply the geometry tracking method detailed in \cref{sec:hrgeo} and albedo texture acquisition in \cref{sec:albedotex} to obtain the tracked template mesh and its corresponding unwrapped mean albedo texture for each phone capture.

\subsubsection{Fitting Environment Lighting}
\label{sec:light}
We parameterize the environment illumination at the time of phone capture as $N$ point lights with position and RGB intensity, $\{\rvl^{\text{pos}}_{i}, \rvl^{\text{int}}_{i}\}^{N}_{i=1}$.
In this paper, we use $N=512$ distant point lights that we assume to be uniformly distributed on the sphere.
During fine-tuning, we freeze the parameters of the encoders and decoders, optimizing only the intensities $\{\rvl^{\text{int}}_{i}\}^{N}_{i=1}$ of the lights to minimize 
the L1 loss between the rendered image $I_f$ and the ground truth $\hat{I_f}$ in the face region:
\begin{align}
\gL_{\text{fit-light}} = \left\|I_f - \hat{I_f}\right\|_1 \text{.}
\end{align}
We show the optimized environment lights of our method in \cref{fig:fit-light}. Compared to the data captured by the phone, our refined light accurately reconstructs the primary light source within the space. This adjustment brings the color space of the avatar more in line with the observed data.

\subsubsection{Fine-tuning Encoders and Decoders}
After fitting the environment lights, we freeze them and finetune all the encoder and decoder parameters $\Phi_{\text{id}}$, $\Phi_{\text{exp}}$, $\Phi_{\text{g}}$, $\Phi_{\text{fi}}$, $\Phi_{\text{fv}}$, $\Phi_{\text{eye}}$, $\Phi_{\text{ei}}$, $\Phi_{\text{ea}}$, and $\Phi_{\text{ev}}$.
The loss function employed during the fine-tuning stage mirrors that utilized in the pre-training stage, with the inclusion of two additional terms:
\begin{align}
\mathcal{L}_{\text{finetune}} = \mathcal{L}_{\mathrm{rec}} + \mathcal{L}_{\mathrm{reg}} + \lambda_{\mathrm{kl}} \mathcal{L}_{\mathrm{kl}} + \gL_{\text{alpha}} +  \gL_{\text{vgg}} \text{,}
\end{align}
where the $\gL_\text{alpha}$ represents the alpha loss on the rendered Gaussian and the input images' predicted alpha matting.
Due to the blurriness inherent in the phone capture data, we incorporate the VGG loss \cite{johnson2016perceptual} between the rendered images and ground truth, denoted as $\gL_\text{vgg}$, during the fine-tuning stage to retain more details from the phone capture.

\subsection{Training details}
We follow previous work~\cite{cao2022authentic} and use a network architecture based on U-Net~\cite{ronneberger2015u} for our identity encoder. The ID encoder consists of a series of convolutional networks that encode the identity conditional data into a set of feature maps, which are then added to each layer's output in the decoder.
Please refer to \cref{fig:network_arch} for networks details.

For the pre-training of the universal relightable prior model, we use the Adam optimizer \cite{kingma2014adam} with a learning rate of $5\times10^{-4}$.
We use 64 NVIDIA A100 GPUs with a batch size of 128 for 400k iterations, which takes 5 days to converge.
For the personalized avatar finetuning stage, we set the learning rate as $10^{-4}$ and finetune the pre-trained networks using 8 NVIDIA A100 GPUs with a batch size of 16 for 13k iterations, where the first 3k iterations are for environment light fitting and the last 10k iterations are for encoder and decoder finetuning.
Including the preprocessing stages, the personalization stage takes approximately 3 hours.

\section{Experiments}
\subsection{Datasets}
\paragraph{Studio Captures.}
Our studio capture setup is similar to the capture system presented by \citet{cao2022authentic} and \citet{saito2024rgca}, where we obtain calibrated and synchronized multi-view images at a resolution of 4,096$\times$2,668\,pixels through the use of 110 cameras and 460 white LED lights operating at 90\,Hz.
Participants are instructed to perform a predetermined set of various facial expressions, sentences, and gaze motions for approximately 144,000 frames.
To observe diverse illumination patterns while maintaining stable facial tracking, we employ time-multiplexed illumination \cite{bi2020deep}.
Specifically, full-on illumination is interleaved every third frame to facilitate tracking, while the remaining frames utilize either grouped or randomly selected sets of 5 lights.
In total, this capture script was used to record data from 345 participants.
Of these, 342 were used to train our universal relightable model, while the remaining three were reserved for evaluation.

\paragraph{Phone Captures.}
We captured 10 people under five alternating illumination conditions via mobile phone. Each video duration is 5–10 minutes per subject. Participants performed a series of actions, including head rotations, changing eye gaze, making various facial expressions, and regular daily speech. 
They were captured in an LED wall cylinder with a diameter of 4.7\,m and a height of 3\,m.
The LED panels have a maximum brightness exceeding 1,500\,cd/m\textsuperscript{2} and a pixel pitch of 2.84\,mm to realistically light the participants under multiple real-world lighting environments.
To estimate the ground-truth environment map for each lighting condition, we first geometrically calibrate the pose of all LED panels, so that we can ray-trace environment maps at arbitrary positions in the capture volume, including the head's center of mass.
Additionally, we characterize the LED panels' response using a Ricoh Theta Z1 360° camera placed at the center of the capture volume, such that we can predict how a particular displayed lighting environment affects a participant in terms of incident illumination.
For this, we display a ramp of red, green, blue and white lighting environments, one at a time, e.g. from 0\% red (black) to 100\% red, and capture a linear HDR image for each condition using a five-photo exposure bracket of raw images. The capture with white LED light is used only for validation.
We further optimize a learnable per-camera 3$\times$3 color correction matrix to account for differences in color spaces between cameras.
For evaluation, we select one environment as input to build a personalized avatar for each subject.
Note that we do not use the ground-truth illumination for our avatar reconstruction.
We then relight the reconstructed avatar in other environments, and compare our rendered avatar with the ground-truth data.
We randomly select 100 frames to cover sufficient variation of head rotation and expressions.

\begin{figure}
  \centering
  \includegraphics[width=\linewidth]{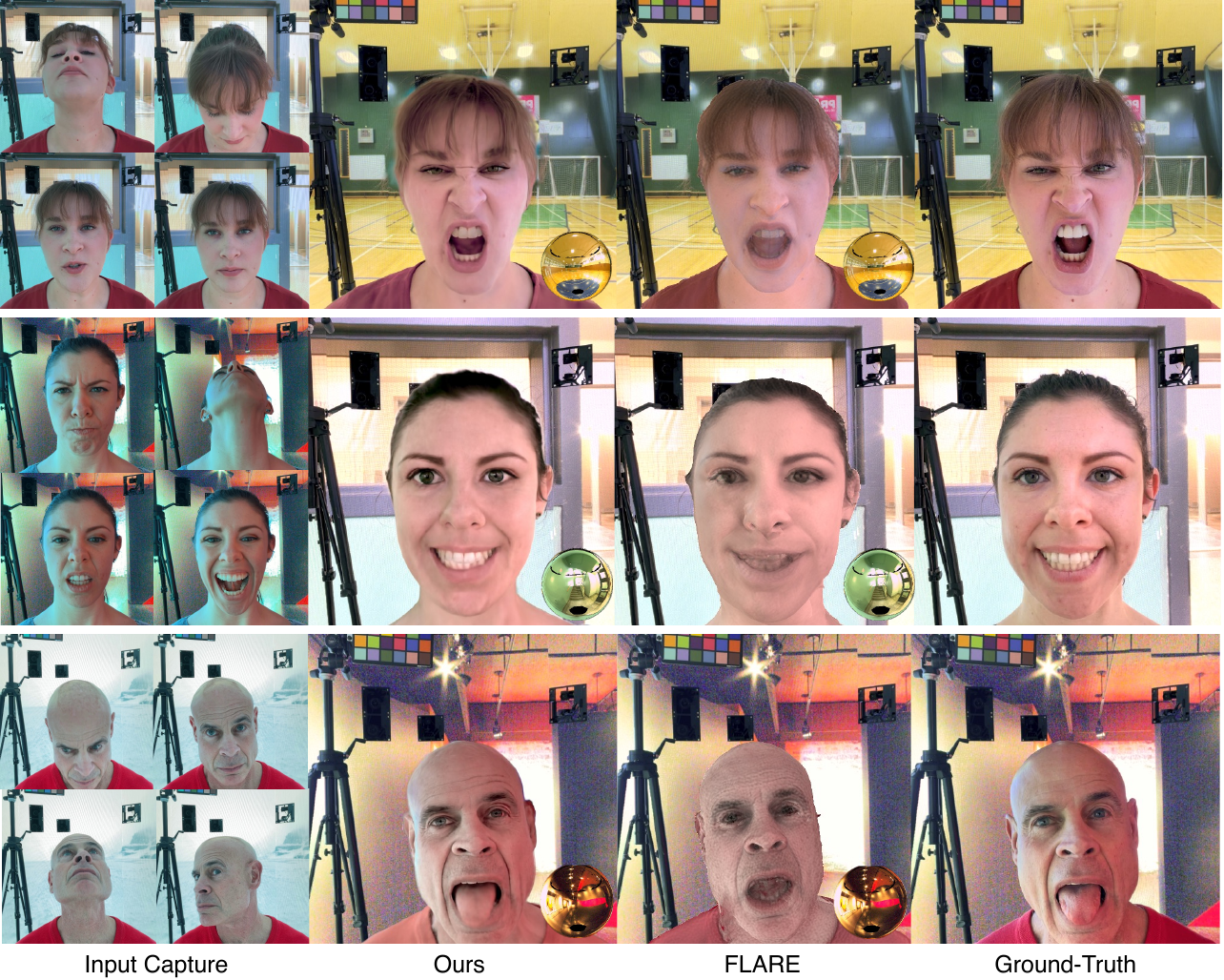}
  \caption{Qualitative comparison of environment relighting between FLARE \cite{bharadwaj2023flare} and our approach in an unseen test environment.}
  \label{fig:env}
\end{figure}

\subsection{Comparisons}
\input{tables/env-relight}
In \cref{tab:env-relight}, we show the quantitative evaluation of our method compared to FLARE, the current state-of-the-art relightable avatar reconstruction method from a phone scan \cite{bharadwaj2023flare}. We report mean-absolute-error (MAE), mean-squared-error (MSE), SSIM and LPIPS as the metrics for the face region.
Unlike our approach, FLARE uses meshes as a shape representation and a parametric BRDF model (i.e., Lambertian for diffuse, and \citet{cook1982reflectance} for specular) with only the first bounce considered.
Our approach achieves significantly lower error, and the qualitative comparison in \cref{fig:env} further confirms the substantial quality improvement.
This illustrates the efficacy of our geometry and appearance representation as well as our universal relightable prior model.

\subsection{Evaluation}

\begin{figure}
  \centering
  \includegraphics[width=\linewidth]{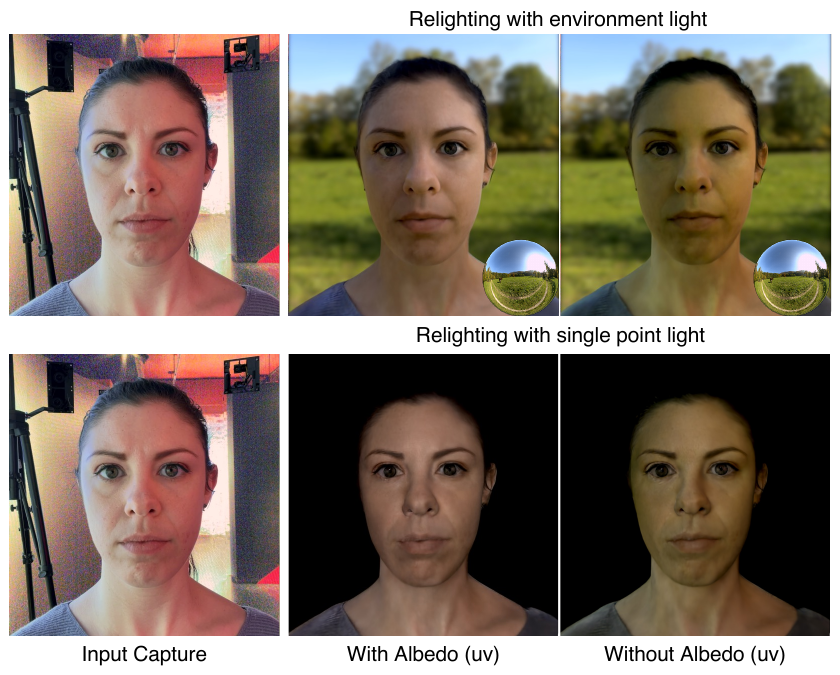}
  \caption{Comparison of w/ vs. w/o using albedo conditioning data.}
  \label{fig:albedo-condition}
  \vspace{-3mm}
\end{figure}

\begin{figure*}
\centering
\includegraphics[width=0.98\linewidth]{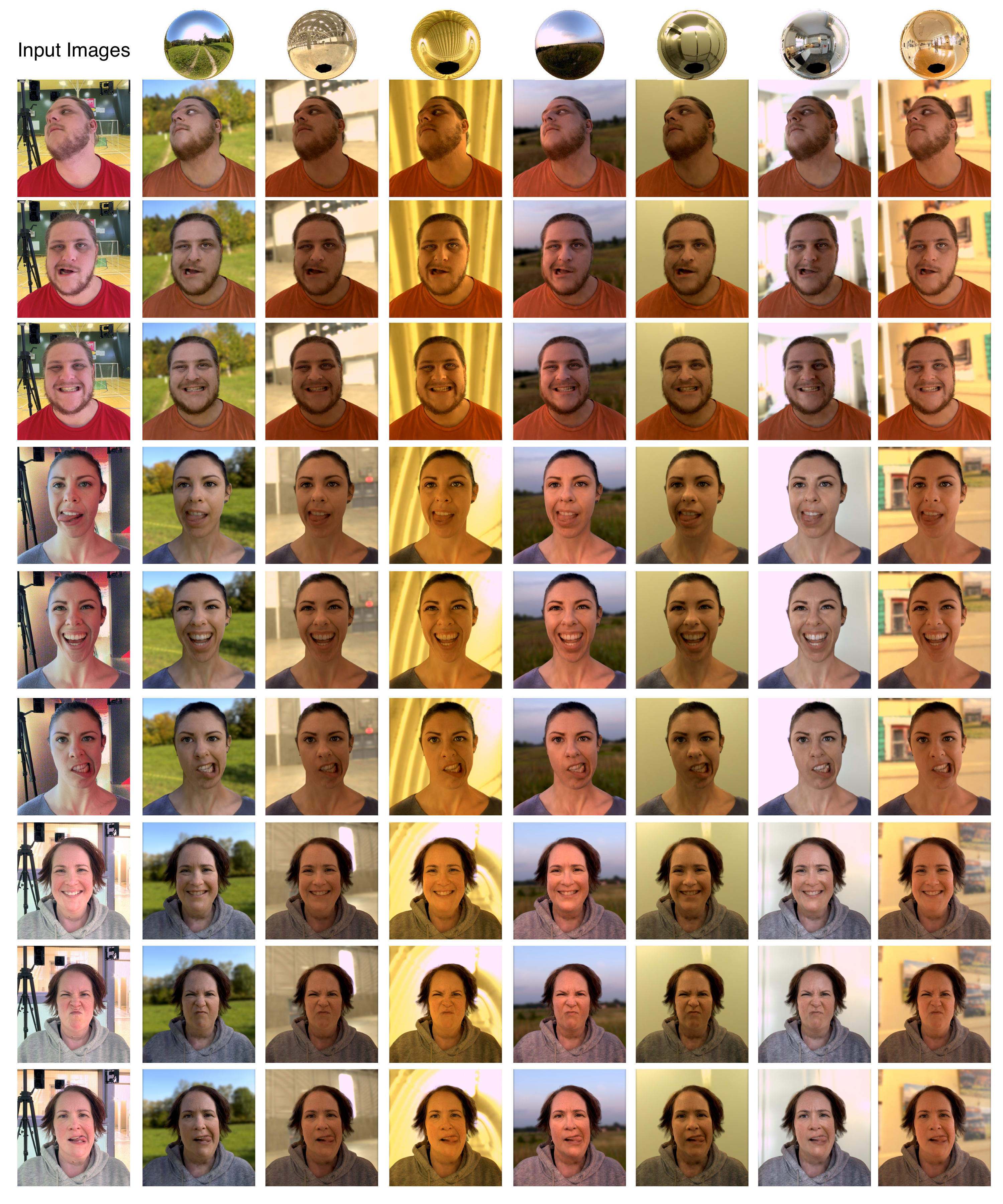}
\caption{Relighting results on multiple identities with different expressions.}
\label{fig:expression_lights}
\end{figure*}

We evaluate our key design choice of using albedo as an identity feature instead of a simple color-transformed mean texture \cite{cao2022authentic}.
\Cref{fig:albedo-condition} shows that illumination-agnostic albedo greatly reduces artifacts in the zero-shot reconstruction case.
In contrast, the simple color transformation leads to more ``baked-in'' artifacts in the reconstructed avatar.
We also conducted an ablation study on the unified eye specular visibility decoder, and the results are shown in \cref{fig:eye_glint_compare}. While the baseline model loses glints due to overfitting to inaccurate ground-truth gaze, our unified decoder faithfully preserve eye glints.
Furthermore, \cref{fig:pica_compare} illustrates the efficacy of the proposed high-resolution tracked mesh over a coarse mesh tracking used in prior work~\cite{cao2022authentic,saito2024rgca} on a novel identity. 
We also show additional qualitative results in \cref{fig:expression_lights}. This shows that our approach generalizes to a wide range of identities, illuminations, and expressions.

\section{Discussion and Conclusion}
The relighting quality of our work could be degraded in several cases.
Our model relies on the universal prior learned from studio-captured data, and variations not covered by the training corpus may lead to suboptimal generalization.
For example, as the captured subjects in the training data all wear gray T-shirts, the relighting of clothing is less accurate than for the head area (see the bottom row in \cref{fig:env}).
Also, while our light estimation is robust, any inaccuracy in the light estimation leads to `baked-in' artifacts.
In particular, estimating high-frequency illumination remains a challenge, which mainly affects the quality of eye relighting.
Future work could address this by incorporating a strong illumination prior \cite{lyu2023dpi,gardner2022rotation}.
Last but not least, the current personalization process requires multiple preprocessing steps and test-time finetuning, and is hence not instant.
Enabling instant personalization with full relighting capability remains exciting future work.

We presented URAvatar, a novel framework to create photorealistic relightable head avatars driven by gaze, neck rotation, and a latent facial expression code in real-time, with faithful diffuse scattering and specular highlights.
Our experiments show that building a strong generalizable prior of global light transport for dynamic facial expressions from multi-identity studio data is now possible with a sufficiently expressive geometric representation (3D Gaussians) and appearance representation (learnable radiance transfer).
We also evaluated the fidelity of our reconstruction on unseen natural illuminations by building a new virtual environment capture system consisting of LED display panels, and showed the efficacy of our method compared to prior works.
Our work, for the first time, enables the faithful reconstruction of relightable avatars with global light transport from a single phone scan, unlocking the possibility of virtual teleportation with authentic avatars as a communication tool.

\bibliographystyle{ACM-Reference-Format}
\bibliography{ref}

%\appendix
%Lorem ipsum dolor sit amet, consectetur adipiscing elit. Morbi
%malesuada, quam in pulvinar varius, metus nunc fermentum urna, id
%sollicitudin purus odio sit amet enim. Aliquam ullamcorper eu ipsum
%vel mollis. Curabitur quis dictum nisl. Phasellus vel semper risus, et
%lacinia dolor. Integer ultricies commodo sem nec semper.
%Etiam commodo feugiat nisl pulvinar pellentesque. Etiam auctor sodales
%ligula, non varius nibh pulvinar semper. Suspendisse nec lectus non
%ipsum convallis congue hendrerit vitae sapien. Donec at laoreet
%eros. Vivamus non purus placerat, scelerisque diam eu, cursus
%ante. Etiam aliquam tortor auctor efficitur mattis.
%Nam id fermentum dui. Suspendisse sagittis tortor a nulla mollis, in
%pulvinar ex pretium. Sed interdum orci quis metus euismod, et sagittis
%enim maximus. Vestibulum gravida massa ut felis suscipit
%congue. Quisque mattis elit a risus ultrices commodo venenatis eget
%dui. Etiam sagittis eleifend elementum.
%
%Nam interdum magna at lectus dignissim, ac dignissim lorem
%rhoncus. Maecenas eu arcu ac neque placerat aliquam. Nunc pulvinar
%massa et mattis lacinia.

\end{document}

%% file: math_commands.tex
\usepackage{amsmath,amsfonts,bm}

\def\eqref#1{equation~\ref{#1}}
\def\1{\bm{1}}

\def\rvl{{\mathbf{l}}}

\def\rvn{{\mathbf{n}}}

\def\rvr{{\mathbf{r}}}

\def\rvt{{\mathbf{t}}}

\def\rvz{{\mathbf{z}}}

\def\mG{{\bm{G}}}

\def\mT{{\bm{T}}}

\DeclareMathAlphabet{\mathsfit}{\encodingdefault}{\sfdefault}{m}{sl}
\SetMathAlphabet{\mathsfit}{bold}{\encodingdefault}{\sfdefault}{bx}{n}

\def\gD{{\mathcal{D}}}
\def\gE{{\mathcal{E}}}

\def\gL{{\mathcal{L}}}

\def\gN{{\mathcal{N}}}

\newcommand{\R}{\mathbb{R}}

%% file: tables/env-relight.tex
\begin{table}[t]
\centering
\caption{Comparison of environment relighting for Phone captured identities. 
}
\label{tab:env-relight}
\begin{tabular}{lcccc}
\toprule
Method      & MAE $\downarrow$ & MSE $\downarrow$ & SSIM $\uparrow$  & LPIPS $\downarrow$ \\ \midrule
FLARE       & 0.0327     & 0.0068    & 0.8849 & 0.1722 \\ 
Ours        & \textbf{0.0136}   & \textbf{0.0025}   & \textbf{0.9524} & \textbf{0.0605} \\
\bottomrule
\end{tabular}
\end{table}